\documentclass{article}

\usepackage{arxiv}

\usepackage[utf8]{inputenc} % allow utf-8 input
\usepackage[T1]{fontenc}    % use 8-bit T1 fonts
\usepackage{hyperref}       % hyperlinks
\usepackage{url}            % simple URL typesetting
\usepackage{booktabs}       % professional-quality tables
\usepackage{amsfonts}       % blackboard math symbols
\usepackage{nicefrac}       % compact symbols for 1/2, etc.
\usepackage{microtype}      % microtypography
\usepackage{lipsum}
\usepackage{amsmath}

\usepackage{subcaption}
\usepackage{rotating}
\usepackage{tikz}
\usepackage{lscape}
\usetikzlibrary{positioning}
\usepackage{array,calc}
\newlength\lena \newlength\lenb \newlength\lenc \newlength\lend
\newcolumntype{P}[1]{>{\centering\arraybackslash}p{#1}} % centered "p" column
\title{Composite Travel Generative Adversarial Networks for Tabular and Sequential Population Synthesis}

\author{
  Godwin Badu-Marf \\
  TRIP Lab\\
  	Department of Geography, Planning and Environment\\
  	Concordia University, Montreal\\ \texttt{godwin.badu-marfo@mail.concordia.ca} \\
   \And
  Bilal Farooq \\
  Laboratory of Innovations in Transportation (LiTrans)\\
  Ryerson University\\
  Toronto, Canada \\
  \texttt{bilal.farooq@ryerson.ca}
  \And
  Zachary Paterson \\
  TRIP Lab\\
  Department of Geography, Planning and Environment\\
  Concordia University, Montreal\\ \texttt{zachary.paterson@concordia.ca} \\
  \\
}

\begin{document}
\maketitle

\begin{abstract}
Agent-based transportation modelling has become the standard to simulate travel behaviour, mobility choices and activity preferences using disaggregate travel demand data for entire populations, data that are not typically readily available. Various methods have been proposed to synthesize population data for this purpose. We present a Composite Travel Generative Adversarial Network (CTGAN), a novel deep generative model to estimate the underlying joint distribution of a population, that is capable of reconstructing composite synthetic agents having tabular (e.g. age and sex) as well as sequential mobility data (e.g. trip trajectory and sequence). The CTGAN model is compared with other recently proposed methods such as the Variational Autoencoders (VAE) method, which has shown success in high dimensional tabular population synthesis. We evaluate the performance of the synthesized outputs based on distribution similarity, multi-variate correlations and spatio-temporal metrics. The results show the consistent and accurate generation of synthetic populations and their tabular and spatially sequential attributes, generated over varying spatial scales and dimensions. 
\end{abstract}

\keywords{Population synthesis, generative adversarial networks, generative models, tabular data, sequential data, microsimulation, agent based modelling}

%\end{frontmatter}


\section{Introduction}\label{intro}

Agent-based transportation microsimulation models study the interaction between the mobility of travel agents and how urban systems operate and evolve through an individual's daily activities \cite{tanton2014review, farooq2013simulation, harland2012creating, farooq2012towards, ryan2009population}. These models help to understand and predict future travel demand, which subsequently impacts transportation networks, environmental sustainability, land and energy usage. 

Traditionally, individual level data have been collected through phone surveys, household or individual travel diaries and paper questionnaires administered by Census agencies. The proliferation of pervasive technologies (i.e. smartphones, mobile devices, GPS) with high computing power and data connectivity capacities in recent times have influenced the volume, variety and velocity of travel data collected \cite{badu2019perspective}. While data collection technologies are advancing, the availability of microdata still remains relatively limited owing to the high cost of acquiring reliable data and also the threat to privacy of the collection of spatially- and temporally-detailed information on individuals. In practice, government bodies (e.g. census agencies) conduct travel surveys on a sample of a population whose statistical characteristics are used to represent the behaviour of the entire population. Using sample data and other information (i.e. partial views) as base population information, researchers can reconstruct  representative members of a population using synthesis techniques such as Iterative Proportional Fitting (IPF) \cite{fienberg1981iterative, wilson1976new}, combinatorial optimization (CO) \cite{muller2010population}, or Markov chain Monte Carlo (MCMC) simulation \cite{farooq2013simulation}.

Deep generative models have evolved recently and shown the ability to estimate the joint probability distribution of data using deep neural networks and have had success in regenerating high resolution images \cite{goodfellow2016deep, kingma2013auto, goodfellow2014generative}. Well known deep generative models, such as Variational Auto-Encoders (VAE) \cite{kingma2013auto} and Generative Adversarial Networks (GANs) \cite{goodfellow2014generative} have gained considerable attention recently for their potential to generate synthetic representations from latent space that estimate the underlying data distributions. GANs have exhibited flexibility in generating high-quality synthetic images and natural language processing \cite{collobert2011natural, chowdhury2003natural}. VAEs use a probabilistic graphical formulation of creating models into latent space thus inherently reducing most dimensions into compressed latent representations. This allows VAEs to train efficiently, but their synthetic outputs can be blurry due to drawing from low dimensional latent space. GANs are explicitly optimized for synthetic generation, and don't have the dimension collapse issues of VAEs. The advantage of GANs is in reproducing realistic synthetic outputs using their adversarial objectives. In this paper, we develop GANs models for population synthesis to estimate combinations of high dimensional synthesized output.

While traditional population synthesis techniques are mostly used for the generation of point estimates and cross tabulations of tabular data, travel behaviour data require spatial and temporal sequences of travel-related activities. Deep neural networks such as Recurrent Neural Networks(RNN) and Long Short Term Memory (LSTM) models \cite{hochreiter1997long} have proved successful in generating sequences through modelling the conditional probability distributions of input sequences. Another contribution of our work is to simultaneously recreate the location sequence of a synthesized population using LSTM, while studying the underlying distribution of the trajectory of the sample. To the best of our knowledge, this is the first effort in the population synthesis literature that recreates disaggregate microdata with sequences of locations.

In this paper, we present a novel composite GANs model following the Coupled GANs architecture by \cite{liu2016coupled}, having multiple generative and distributive models to learn the joint distribution of multi-domain travel diary data having tabular socio-economic variables as well as sequential trajectory locations. This model is capable of learning the joint distribution by drawing samples from the marginal distributions of variables. In summary, our contributions expand on the current literature on population synthesis as follows:

\begin{enumerate}
	\item We propose a composite GANs architecture to simultaneously recreate synthetic representations of tabular microdata \textit{and} sequential locations of travel diary data. 
	\item In tabular microdata synthesis, we synthesize mixed features i.e. numerical as well as categorical.
	\item We showcase synthetic sequences of locations inspired by the SeqGAN \cite{yu2017seqgan}.
	\item We compare and evaluate the performance and similarity of synthesized tabular data distributions to synthesis using Variational Autoencoding \cite{borysov2018scalable}.
\end{enumerate}

The paper is organized as follows. In Section 2, the literature review is provided. Section 3 formalizes the problem and introduces the proposed methodology. In Section 4, a case study, evaluation procedure, results and discussion are provided. Section 5 provides a conclusion and some directions for future work.

\section{Literature review}
Traditional population synthesis approaches have been inherently mathematical and can be used to estimate synthetic members of a population having spatial and aspatial characteristics. The aggregate summary of population members corresponds to published aggregates of the entire population. These synthesis approaches are broadly classified into three categories namely, re-weighting, matrix fitting, and simulation-based approaches \cite{tanton2014review}. First of all, re-weighting methods adopt different techniques to adjust the weight factor of surveys such that the sample represents subregions rather than the entire summation of the population aggregates. In this sense, re-weighting applies non-linear optimization to estimate weights and are not scalable to high dimensions \cite{bar2009estimating, daly1998prototypical, harland2012creating}. Matrix fitting method evoke expansion factors that are expressed by the ratio between a starting solution and the final matrix. Common implementations of matrix fitting are the Iterative Proportion Fitting (IPF) proposed by \cite{deming1940least} and the Maximum Cross-Entropy \cite{guo2007population}. It is worth noting that these two methods known as deterministic models, do not produce agent-based samples but rather a sample of prototypically weighted agents \cite{borysov2018scalable}. Lastly, simulation-based approaches model the joint distribution of population data with its full set of attributes. New members of the population can be recreated by sampling from the joint distribution. This approach addresses the drawbacks of the deterministic models and is capable of estimating agent-based samples while being scalable to high dimensional datasets. A notable simulation-based approach is the Bayesian Network proposed by Sun and Erath \cite{sun2015bayesian}. This method uses a graphical representation of a joint probability distribution, encoding probabilistic relationships among a set of variables in an efficient way. While the bayesian network outperforms the deterministic models, the learning of its graph structure can be computationally challenging \cite{borysov2018scalable}.

More recently, deep generative models have become popular in the academic literature because of their outstanding performance and computational effectiveness in producing realistic images and machine translation \cite{salimans2016improved, berthelot2017began}. Well-known deep generative models are the Variational Autoencoder (VAE) \cite{kingma2013auto}, restricted Boltzmann Machines (RBM) \cite{wong2020bi}, and Generative Adversarial Networks (GANs) \cite{goodfellow2014generative}. These generative models have shown promising results in reproducing the structural and statistical representations of original data by sampling from the estimated joint probability distribution of the underlying data. While GANs have been used extensively for image, sound and sequential text generation, little attention has been paid to its applications in terms of structured tabular data that is mostly composed of numerical and categorical features.

\cite{choi2017generating} proposed a model that combines auto-encoders with GANs to synthesize private electronic health records. Their method focused on the generation of binary and count variables in health datasets. The authors assert that the original ``vanilla" GANs formulation \cite{goodfellow2014generative} is susceptible to the ``mode collapse" problem and difficult to train \cite{salimans2016improved}. Similar work by \cite{park2018data} proposed a \textit{table-GAN} to synthesize tabular data using a hinge-loss privacy control mechanism. Their method showed a compatible model for anonymization as sensitive attributes are maintained without change. Recently, Borysov et al. \cite{borysov2019scalable} presented a generative model to synthesize micro-agents from a large Danish travel diary to learn the joint distribution of the training data using a Variational Autoencoder (VAE) model. In our approach, the GANs architecture will be optimized for high performance throughput, making it capable of learning all training data records; even those with many zeros representing agents that are omitted from the training samples but exist in the real population.

Generative models have been used in the generation of sequence discrete data, such as text and language translation. Sequence prediction models are typically trained to maximize the log-likelihood (Maximum Likelihood Estimation, or MLE) of the next token (character or word) based on the current token. GANs has had little progress in generating sequence discrete data \cite{huszar2015not} because the generator network is designed to output continuous gradient updates, which does not work on discrete data generation \cite{goodfellow2016deep}. In an attempt to solve this discrepancy, Bengio et al.\cite{bengio2015scheduled} proposes Scheduled Sampling builds on MLE by randomly replacing ground-truth tokens with model predictions as the input for decoding the next-step token. Another approach is to use the concept of Reinforcement Learning named SeqGAN  \cite{yu2017seqgan}. The SeqGAN approach models the generator as a stochastic policy where the state is the tokens generated so far and the action is the next token to be generated. The presence of a stochastic policy, REINFORCE \cite{ranzato2015sequence} algorithm, allows different actions to be sampled during training and derive a robust estimate of the policy. We adopt the SeqGAN approach in our model for the sequential component of the CTGAN architecture whose purpose is to synthesize trip sequences.

\section{Methodology}
The problem definition is introduced, which establishes the objective of this research. As a base case, we briefly present the variational auto-encoder method, which has been recently used for population synthesis of tabular data only \cite{borysov2018scalable}. An overview of the Generative Adversarial Networks and subsequently a detailed description of our proposed composite architecture of GANs for synthesizing tabular and location sequences follow.

\subsection{Problem definition}
We assume a dataset on mobility of \textit{N} population agents (i.e. households, families or individuals) characterized by a set of basic attributes $X=(X_1, X_2, X_3, ... X_m)$ where \textit{m} is the number of attributes, and their sequence of time-ordered trips to locations drawn from the universe of locations, \textit{U}$_L$. The universe of locations, without loss of generality, consists of geographic positions of all route intersections and road vertices within the study area. Formally, the trip chain is defined by $T = L_1 \rightarrow L_2 \rightarrow \cdots \rightarrow L_{|T|}$ where $\forall 1 \leq i \leq |T|, L{i} \in \textit{U}{_L}$. It is worth noting a location may occur multiple times in a sequence of trip chain especially for home based trips. Table \ref{tab:my-table} shows an example of such a dataset. Typically, the joint distribution between attributes in a true population are not accessible hence partial views such as samples are used to estimate the joint distribution of the population \cite{farooq2013simulation}. In this regard, we present a novel generative framework using deep learning methods to estimate the joint distribution of a true population using sample partial views having tabular and sequential attributes, from which we can draw synthetic agents with tabular and sequential characteristics simultaneously.

\begin{table}[!hb]
	\small
	\centering
	\begin{tabular}{|l|l|l|l|l|}
		\hline
		\textbf{Age (x$_1$)} & \textbf{Sex (x$_2$)} & \textbf{Status (x$_3$)} & \textbf{Permit (x$_4$)} & \textbf{Trips (T)}      \\ \hline
		21           & m            & student       & y      & $ L1 \rightarrow L2 \rightarrow L3 \rightarrow L4 $ \\ \hline
		30           & f            & worker        &n         & $ L1  \rightarrow L3 \rightarrow L4 $ \\ \hline
		45           & m            & not employed      &y       & $ L1 \rightarrow L2 \rightarrow L3 \rightarrow L4 $ \\ \hline
	\end{tabular}
	
	\caption{A preview of mobility data on travel agents comprising structured and sequential features.}
	\label{tab:my-table}
\end{table}

\subsection{Variational Auto-encoders}
The Variational Auto-Encoder (VAE) was proposed by \cite{kingma2013auto}, as an alternate deep learning approach to estimate a population distribution into a compressed lower dimensional latent space using a neural network called the ``encoder" that is supported by an auxiliary neural network named the ``decoder", acting as a generator by drawing random samples from the distribution of the latent space. During training, the encoder network receives an input vector of the size of the training data and outputs a latent representation. The decoder network receives the latent representation as input and generates new synthetic agents from the prior distribution of the latent space. Using VAE, \cite{borysov2018scalable} developed a scalabe population synthesis method for tabular data and showed that it outperforms IPF and simulation based methods. Thus we will use VAE as our base case for comparison for tabular data synthesis (Columns 1--4 in Table \ref{tab:my-table}). For further reading about the VAE, readers are referred to \cite{kingma2013auto, borysov2018scalable}.

\subsection{Generative Adversarial Networks}
\cite{goodfellow2014generative} proposed Generative Adversarial Networks (GANs), which have gained prominence in the deep learning literature because generative modelling has shown promising results in synthesizing realistic images and sequences for natural language processing. Intuitively, GANs simulates a two player game composed of Generator and Discriminator networks. The goal of the Generator is to generate samples from latent space that are equivalent to real samples while the Discriminator acts as a police officer to distinguish real samples from synthesized ones. Models of the generative and discriminator are both realized as multilayer perceptrons. During model learning, the Discriminator gets better at discriminating real samples from fake, while the generator improves on generating samples that are close to the real samples until a Nash equilibrium \cite{nash1950equilibrium, lim2017geometric} is achieved, where each model reaches its peak ability to thwart the other's goal. The objective function of GANs is defined as:

\begin{flushleft}
	
	\textbf{Definition 1 (Objective function): }
	
	The \textit{objective function} of the Generative Adversarial Networks \cite{goodfellow2014generative} is:
	
	\begin{equation} \label{eq1}
	\displaystyle {G}_{min} \ {D}_{max} \ V(D,G)= \mathbb{E}{_{x \sim p_{data}(x)}}[log \ D(x)] + \mathbb{E}{_{z \sim p_{z}(z)}}[log(1 - D(G(z)))] 
	\end{equation}
	
\end{flushleft}

Equation \ref{eq1} explains the objective function of the Discriminator, which seeks to maximize the output of D(x) to 1 when the input is from the true data distribution of the real samples. If the input is generated from the Generator, then D(G(z)) should minimize the output of the objective function. In the training process, both networks simultaneously learn parameters using Stochastic Gradient Descent. The training process halts when a Nash equilibrium is reached so that the Discriminator is unable to distinguish probability from true or fake samples.

\subsection{Coupled generative adversarial network}
The Coupled generative adversarial network (CoGAN) proposed by \cite{liu2016coupled} addresses the problem of learning a joint distribution of multi-domain images from data. While other multi-modal learning approaches exist \cite{srivastava2012multimodal, wang2012semi, ngiam2011multimodal}, CoGAN has shown successes in overcoming correspondence dependency\cite{liu2016coupled} which makes it challenging to build a dataset of corresponding images. CoGAN is built on the GANs framework \cite{goodfellow2014generative} and extends the capability of learning joint image distribution tasks. CoGANs consist of multiple GANs networks each defined for a single image domain. While CoGAN naively learns the marginal distributions of its input data, the authors enforced a weight-sharing constraint to achieve joint distribution learning between the networks and showed its effectiveness in application to multi-image domains, unsupervised domain adaptation and image transformation. We refer readers to the literature \cite{liu2016coupled} for a thorough discussion on the architecture and applications of the CoGAN.

\subsection{Composite Travel Generative Adversarial Network}
\begin{figure}[!h]
	\centering
	\includegraphics[width=\textwidth]{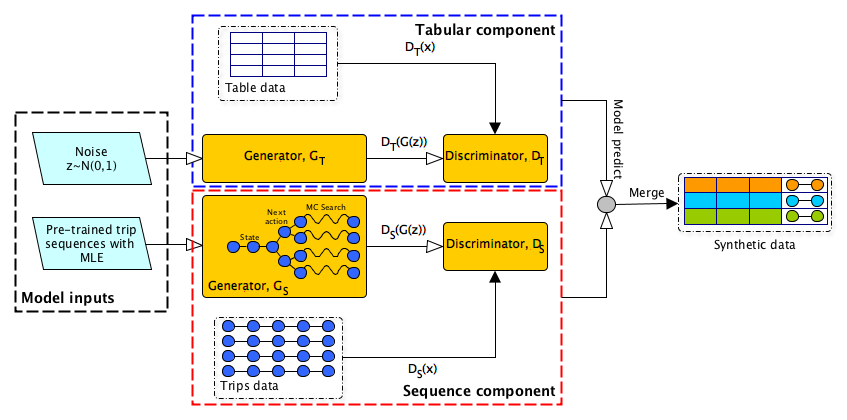}
	\caption{The architecture diagram of Composite Travel Generative Adversarial Networks (CTGAN).}
	\label{fig:ctgan}
\end{figure}

The Composite Travel Generative Adversarial Networks (CTGAN) is designed for learning the joint distribution of tabular travel attributes \textit{and} sequential trip chain locations of an agent in a simultaneous manner, drawing inspiration from the CoGAN proposed by \cite{liu2016coupled}. CTGAN as shown in Figure \ref{fig:ctgan}, consist two GAN networks - GAN$_1$ referred as the Tabular model, and GAN$_2$ as the Sequence model. 

The CTGAN has a \emph{tabular} component whose objective is to learn the joint distribution of the basic socio-demographic attributes in the travel diary and a \emph{sequential} component with an objective to learn the distributions of the trips undertaken in a day by an agent. During the training, each component is implemented as an independent network and learns its parameters based on the underlying data distribution. CTGAN then learns to synthesize pairs of tabular attributes with sequential locations of an agent in a population.

\subsubsection{GAN$_1$-Tabular Component}

The purpose of the \emph{Tabular} component in the CTGAN is to synthesize the table of records on an agent's socio-demographic and economic attributes (i.e. Age, Sex, Status, Income) which exist in numerical as well as categorical types. GAN$_1$ is able to synthesize both types of tabular attributes. 

\begin{figure}[!h]
	\centering
	\includegraphics[width=0.90\textwidth]{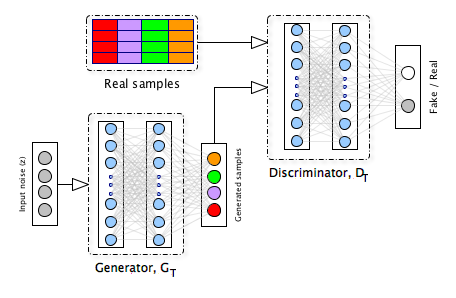}
	\caption{The structure of the Tabular component of CTGAN}
	\label{fig:gan1}
\end{figure}

The \emph{tabular} component shown in Figure \ref{fig:gan1} is composed of an independent GANs architecture having a single Generator denoted G$_T$ and Discriminator, D$_T$. The Generator, G$_T$, is made up of a Multi-Layer Perceptron (MLP) with neurons for each layer connected to the neurons of the next layer. It takes as input a fixed set of vectors and processes them through three (3) hidden layers to compute a higher level representation of the inputs. A final output layer returns a prediction of a last representation for the corresponding inputs. Similar to the Vanilla GAN \cite{goodfellow2014generative} implementation, the input layer of the Generator accepts a random noise sampled from a Gaussian distribution with a dimension size equivalent to the size of the real data. In order the depth of features learnt in the neural network, we exploit multiple hidden layers in the network. Each layer has a bias with a Rectified Linear Unit (ReLU) \cite{hinton2010rectified} activation applied to its output. The ReLU activation is used because it is computationally efficient which allows the network to convert faster, and easily allows for backpropagation. Due to the diverse nature of the data types (i.e. numerical and categorical), the final output layer is split into categorical and numerical vectors. For the categorical vectors, the Softmax activation is applied while the Linear activation function is applied to the numerical vectors. Subsequently, the activated output layers are merged together as a final output of the generator network. We consider age a continuous numerical feature unlike previous work of \cite{borysov2018scalable} that bins into age group categories using count aggregates. An arbitrary size of 200 neurons are defined for the first hidden layer, followed by 100 neurons and 50 neurons for the last hidden layer. The choice of neuron sizes was done randomly and the best choice was based on the training performance of the network and distribution of the final output layer. 

The Discriminator of the tabular component, D$_T$, is designed with an aim to distinguish between true data and synthetic data from the Generator, G$_T$. The Discriminator is made up of Multi-Layer Perceptron with neurons for each layer connected to the neurons of the next layers. The input layer of D$_T$ receives a matrix with the size of the true data shape equivalent to the size of the generated data from G$_T$. The real data samples are pre-processed prior to being fed into the input layer. The numeric features are normalized to a range between -1 and 1, a recommended approach for optimizing effective learning in neural networks \cite{goodfellow2016deep}. The binary and categorical features were encoded with one-hot vectors \cite{guo2016entity} because of the low cardinality of categorical unique values. Each hidden layer is composed of matrix multiplication of nodes with bias and a ReLU activation function. The last hidden layer is activated with a Sigmoid activation function with output of 1 for real samples and 0 for fake samples.

\subsubsection{GAN$_2$-Sequential Component}
\begin{figure}[!h]
	\centering
	\includegraphics[width=0.90\textwidth]{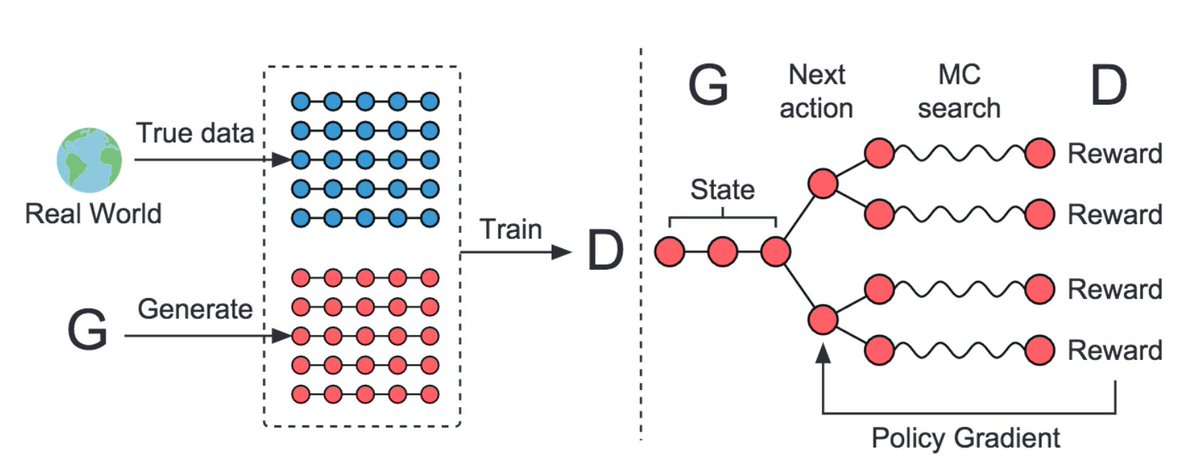}
	\caption{The sequential architecture diagram from SeqGAN \cite{yu2017seqgan}}
	\label{fig:gan2}
\end{figure}

In the second component of the CTGAN architecture, the objective is to synthesize sequences of location distributions traveled by population agents. As earlier mentioned, the CTGAN is composed of multiple generators and discriminator networks hence for the second network of generator and discriminator, we adopt and integrate the SeqGAN model shown in Figure \ref{fig:gan2} proposed by \cite{yu2017seqgan} that has been successful in the generation of text sequences. This network cluster is referred as the ``Sequential component of CTGAN." We extend the implementation of this architecture towards synthesizing location sequences knowing that previous work has used the same in text and sentence generation \cite{hu2017toward, zhang2017adversarial, fedus2018maskgan}. 

It is worth noting that GANs have proven difficulty in the training and generation of sequences and discrete data types. By design, the standard GANs were designed to work with continuous or real-valued data, thus the gradients propagated from the discriminator exist as floating or real-valued losses sent to the generator. This implementation limits the suitability of training with gradient descent on discrete data types. Another is in how the discriminator evaluates gradient loss on a sequence. The discriminator is designed to only classify and evaluate gradient loss on an entire sequence. For instance, only a complete sentence of text can be classified as real or fake by the discriminator but not an incomplete sentence with parts of text. This implies that the loss of a partial sequence cannot be evaluated on how good the partial sequence is until the entire sequence is fully generated.  

This scenario cannot be applied to discrete types as they cannot be updated with continuous or real-valued losses. In order to address the drawback of evaluating partial sequences, we adapt the SeqGAN approach to employ an intermediate score mechanism built using Reinforcement Learning \cite{kaelbling1996reinforcement}. The intuition of Reinforcement learning is illustrated by an agent (a baby) who takes a set of actions (like walking) in an environment based on the state (or thinking) of the agent. When the outcome of the actions of the agent is successful, the agent is given a reward. The objective of this approach is to optimize the actions of the agent and adversely maximize the future expected rewards to the agent. In this regard, the Generator, G$_S$ is modelled as an agent of Reinforcement Learning as discussed. As an RL agent, the state \textbf{s} is defined as the tokens generated so far, the action \textbf{a}, as the next token to be generated and a Reward \textbf{r} gives an intermediate feedback or score to guide G$_S$ by D$_S$ on evaluating the location sequence generated. The gradients from the Discriminator, D$_S$, cannot pass back to G$_S$ since the outputs are discret. To overcome this, we implement an algorithm of reinforcement learning called ``Policy Gradient" which is a stochastic parameterized policy. As a stochastic parameterized policy, the action (next token) may be sampled from a normal distribution whose parameters (i.e., mean and variance) are predicted by the policy. When the samples drawn are evaluated by the policy, subsequent samples can be drawn by moving mean closer to samples that lead to higher rewards, or farther away to samples leading to lower reward. The underlying objective of the generator model is to generate a sequence starting from a state $S_O$ in a way to maximize the expected end reward. We discuss the definition of the end reward in the next section.   

\begin{flushleft}
	\textbf{Definition 2 (End Reward):} 
	
	The \textit{expectation of the end reward} is defined by:
\end{flushleft}

\begin{equation} \label{eq3}
J(\theta)=\mathbb{E}[R_T|s_0, \theta] = \sum_{y_1\epsilon Y}G(y_1|s_0).Q_{D}^{G}(s_0, y_1)
\end{equation}

The expectation of the end reward R$_T$ given in Equation \ref{eq3} is derived as the product of possible values of the reward (i.e. the action-value function) and the probability of the value occurring when given a start state s$_0$, and Generator with parameter of $\theta$. The action-value function $Q_{D}^{G}(s_0, y_1)$ estimated by the discriminator returns the reward value for taking an action from the state $s_0$ following the policy G. The objective of the Generator, G$_S$ is to generate sequences of location destinations from the start state \textbf{s$_0$} in a way to maximize the end reward, \textbf{R$_T$} determined by D$_S$. While D$_T$ only rewards the end of a finished sequence, it is important for every action predicted at each timestep of a state be evaluated for fitness. Intermediate scores are thus required. To achieve this, the Monte Carlo search with roll-out policy is used as in SeqGAN. This approach samples the unknown tokens and estimates the state-action value at each intermediate step. 

The Monte Carlo search is a tree-search algorithm having a root node, \textbf{s$_0$}. The root node is expanded while trying all possible actions belonging to the set of action states as a way to construct child nodes for each state. The value for each child node is determined while the remaining tokens are rolled out with a policy until the entire sequence is generated. The Discriminator gives a score accumulated on each node of the MC tree when the end of sequence is reached. 

\section{Data and case study}

The experimental evaluation of CTGAN is based on travel data from the 2013 Montreal Origin-Destination (OD) survey conducted in 2013. The data contains the travel diary of 139,354 individuals and includes socio-economic variables such as age, employment status, gender, etc., and other trip related variables such as origin and sequence of trip destinations \cite{amt2013enquete}.

\subsection{Data Pre-processing}
Dealing with the mixed data types and complex geospatial types, especially for generating travel survey data poses two challenges: numeric representation, and reversibility. Neural networks work efficiently with floating precision numbers, making it necessary to translate all variables into low-cardinal dimension floating representations and to ensure the uniqueness of each sample represented. 
Binary and categorical variables are indexed numerically and one-hot encoded \cite{potdar2017comparative}. Numeric variables are scaled and normalized within a range from negative one (-1) and positive one (+1). These pre-processing techniques derive a numeric representation of the input data. Unlike regression and classification algorithms that usually have a single output, generative modelling of tabular data requires the vectors of the final output layer to be easily reversible to readable formats synonymous with the raw input data. Thus, encoding techniques of input data to numeric representations must be easily reversible with the ability to be decoded to the format of the input data. In our work, we used Scikit-Learn \cite{bisong2019introduction} label encoding and OneHot encoders which have reverse encoding capabilities. The geographic coordinates (i.e. latitudes and longitudes) of spatial locations are transformed into one-dimensional spatial representation using the Google s2 \cite{googleS2geometry} library. The travel routes were generated using the shortest distance path between origin and destination points. This was implemented using the Open Source Routing Machine (OSRM) api available at http://project-osrm.org. 

\section{Evaluation metrics and results}

We evaluate the fitness of the synthesized population using similarity benchmarks on the statistical and spatial distribution. As a base case for comparison we also synthesized a population using VAE with the same input data. 

\subsection{Similarity in statistical distribution}
The purpose of this benchmark is to evaluate the statistical similarity between the true and synthetic representations of the data. An efficient approach to guarantee the utility of synthetic reconstruction is to compare its statistical properties to the true distribution whose results should be identical or near-identical. We assume that the synthetic data is fit for microsimulation estimations when aggregate queries on both true and synthetic distributions are equivalent. We evaluate the similarity of statistical properties using three (3) metrics. First, we observe the full joint distribution of all possible combinations of data variables. While this approach is efficient for low dimensional tabular data as used in this paper, an implementation to high dimensional data could be complicated. Partial and conditional joint distributions should be used in such cases. Secondly, we derive and compare the marginal distributions for all domains in data variables for the true and synthetic representations. Using this benchmark, the success of the synthesized output is measured by the high score in similarity of the probabilities of values of variables in both datasets without reference to the values of other variables. Finally, we quantify the empirical distributions between the synthetic and true distributions with the Standard Root Mean Square Error (SRMSE) \cite{kirill2011population}, the accuracy and fitness of the synthetic reconstruction using a measure of the Pearson correlation coefficient(\textit{corr}) and the coefficient of determination(\textit{R$^2$}). The standardized root mean squared error is defined by:

\begin{equation} \label{eq4}
SRMSE(\hat{\pi},\pi)=\frac{RMSE(\hat{\pi},\pi)}{\bar{\pi}}=\frac{\sqrt{\sum _i \cdots \sum _j(\hat{\pi} _i... _j -  \pi _i... _j)^2/N _b}}{\sum _i ... \sum _j \pi _{i...j}/N _b}
\end{equation}
where \textit{N$_b$} is the total number of agents; \textit{R$_{i..j}$} is the number of agents with attribute values i...j in the synthesized population, $\hat{\pi}$ and $\pi$  is the synthetic and true distribution respectively.

\subsection{Similarity in spatial distribution}
To evaluate the utility of the synthetic reconstruction on sequential location data, we evaluate with metrics: \textit{trip length, segment usage} and \textit{origin-destination distribution}. Trip length distribution measures the similarity in distances traversed on trip segments, segment usage distribution measures the frequency of trips on a routes and the origin-destination measures the agent count on each zone for trip origin and destinations. These metrics quantify the accuracy and fitness of spatial characteristics in the synthesis model.

\section{Experiments and evaluation results}
In this section, we discuss the experiment setup and the results achieved on the model implementation using the metrics stated. The model was built and implemented with Python Keras with Tensorflow backend support on a MacBook Intel Core i5-4258U and GPU Intel Iris Graphics 5100.

\subsection{Statistical distribution comparison}
In this experiment, we focus on comparisons of population-synthesis-based approaches on tabular data between CTGAN and VAE. The experiments were designed such that both models were provided with the same amount of data and dimensions about the sample population. The output of each model is subsequently analyzed to evaluate how good the full joint and marginals of the true population are reproduced. To assess the goodness of fit, the Standardized Root Mean Square Error is performed on the output of each model.

\begin{figure}%
	\centering
		\begin{subfigure}[t]{.45\textwidth}
		%\centering
		\includegraphics[width=7.3cm]{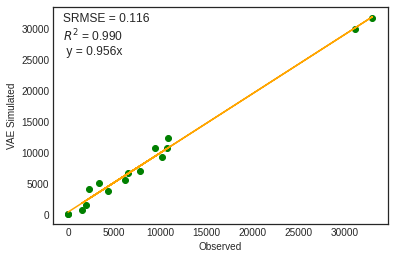}
		\caption{VAE.}
	\end{subfigure}	
			\begin{subfigure}[t]{.45\textwidth}
		%\centering
		\includegraphics[width=7.3cm]{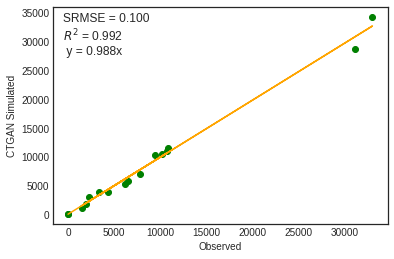}
		\caption{CTGAN.}
	\end{subfigure}	
	\caption{Fit between true and synthesized population.}%
	\label{fig:joint_compare}%
\end{figure}

\begin{figure}[h!]
	\centering
	\begin{subfigure}[t]{.45\textwidth}
	%\centering
	\includegraphics[width=7cm]{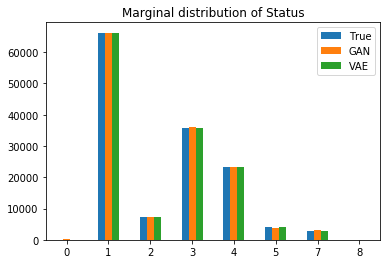}
	\caption{Employment status.}
\end{subfigure}%
	\begin{subfigure}[t]{.45\textwidth}
	%\centering
	\includegraphics[width=7cm]{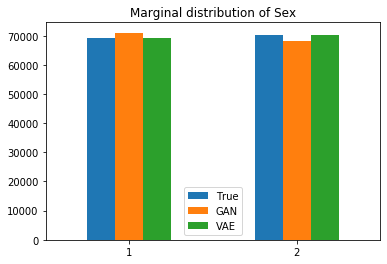}
	\caption{Sex.}
\end{subfigure}

	\begin{subfigure}[t]{.45\textwidth}
	%\centering
	\includegraphics[width=7cm]{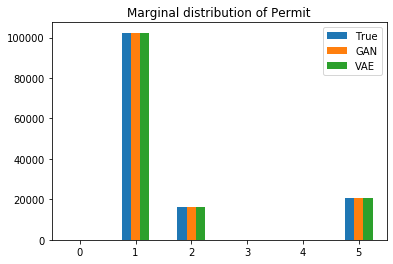}
	\caption{Permit.}
\end{subfigure}%
	\begin{subfigure}[t]{.45\textwidth}
	%\centering
	\includegraphics[width=7cm]{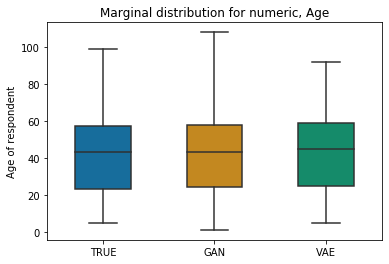} 
	\caption{Age.}
\end{subfigure}	

	\caption{Comparison of marginals for attributes for True, CTGAN and VAE data.}%
	\label{fig:marginal}%
\end{figure}

For comparative analysis on the full joint distribution, we consider a combination of all attributes in the sample data for Age Group (the age variable is discretized into groups of child, young, adult, old), Sex, Employment status and Permit. We construct a contingency table on all combinations of attributes using frequency counts. As observed in Figure \ref{fig:joint_compare}, while both models give a good synthetic representation of the true data distribution, the simulated observations from CTGAN exhibit a better fit with a lower SRMSE of 0.010 while the VAE results in an SRMSE of 0.116. Also, CTGAN results in a strong correlation 0.996 compared to 0.988 for the VAE. The minimal loss in approximation of the VAE could be attributed to the low latent dimensional representation adopted by the VAE thus there is a loss of resolution in the synthetic reconstruction. Similarly as can be seen in Figure \ref{fig:joint_compare}, the VAE shows a slight dispersion along the line of fit that could be attributed to the same low representation. 

The marginal distributions of the tabular variables are shown in Figure \ref{fig:marginal}, and depict the similarity of representation for both the VAE and CTGAN approaches to the True distribution. Obviously, the synthetic population perfectly reproduces the marginals of the training data. The representation from the VAE marginal distribution gives a better similarity to the true distribution than the CTGAN though the model does not memorize the input data. This could be a cost of vanishing gradients suffered by the use of sigmoid activation functions \cite{hochreiter1998vanishing, hanin2018neural} on the last output layer of the generator network for binary types, as seen by the slight imbalance in the marginals of sex variable.

\begin{figure}[h!]
	\centering
			\begin{subfigure}[t]{.45\textwidth}
		%\centering
		\includegraphics[width=7cm]{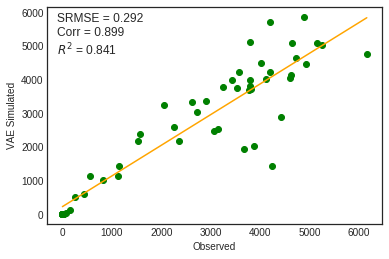} 
		\caption{VAE.}
	\end{subfigure}
			\begin{subfigure}[t]{.45\textwidth}
	%\centering
	\includegraphics[width=7cm]{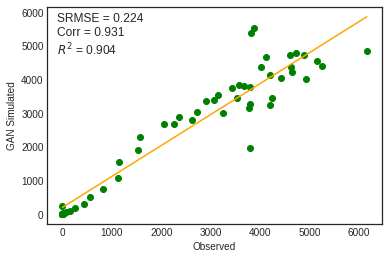}
	\caption{CTGAN.}
\end{subfigure}
	\caption{Fitting and correlational analysis for marginal distribution on numeric variable, Age.}%
	\label{fig:corrAge}%
\end{figure}

We extend the experiment to compare the fitting and correlation patterns in the marginal distributions of the numeric variable, age. As shown in Figure \ref{fig:corrAge}, CTGAN exhibits a better fit with a lower SRMSE of 0.224 compared to SRMSE of 0.292 of the VAE. At an R$^2$ of 90\%, the CTGAN model explains the true distribution with minimum variation relative to the 84\% of the VAE. Finally, it is evident that the simulated agents of the VAE show spread along the best line of fit while agents remain clustered along the line of fit for the CTGAN. In this sense, the CTGAN model presents a reliable agent representation that has a better fit to the true distribution and clearly outperforms the VAE.

\subsection{Spatial distribution comparison}
In order to ensure the consistency in the spatio-temporal behaviour of synthetic agents is retained after synthetic reconstruction of the trip sequences, we evaluate the similarity in trip length distributions and the spatial distributions of error in route segment usage.

\begin{figure}[h!]
	\centering
		\begin{subfigure}[t]{.45\textwidth}
		%\centering
		\includegraphics[width=7.2cm]{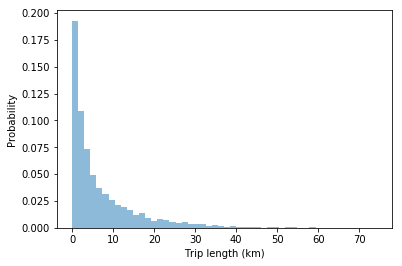}
		\caption{Trip length distribution of True data.}
	\end{subfigure}
		\begin{subfigure}[t]{.45\textwidth}
	%\centering
	\includegraphics[width=7.2cm]{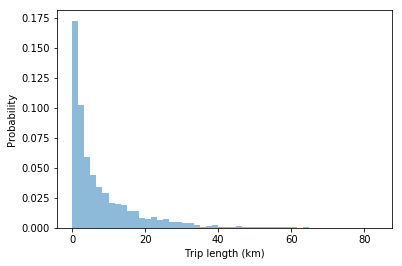}
	\caption{Trip length distribution of Synthetic data.}
\end{subfigure}
			\begin{subfigure}[t]{.45\textwidth}
		%\centering
		\includegraphics[width=7.2cm]{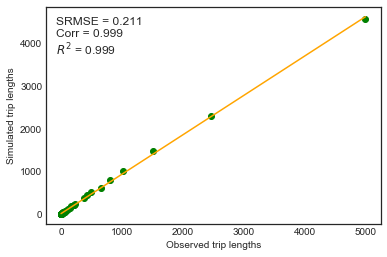} 
		\caption{Line of fit for trip length counts.}
	\end{subfigure}

	\caption{Histogram of trip length distributions for true (a) and synthetic (b), and best line fitting for true and synthetic trip lengths.}%
	\label{fig:trip_len}%
\end{figure}

\subsubsection{Trip length distribution}
Trip lengths are defined by the movement of an agent from one location (origin) to another geographic location (destination). The length of trips is estimated using the euclidean distances between two points. Typically, an agent embarks on a sequence of trips (i.e. trip segments) based on the purpose at the time of the day until a complete trip ends at the start origin. We consider the lengths of all trip segments and compare the frequency distribution of travel distances between the true and synthetic sequential representations.

In Figure \ref{fig:trip_len}a and \ref{fig:trip_len}b, the CTGAN simulated trip lengths show a near equivalence in distribution to the real sequences. It is observed that there is a high count of short trips within distances of two (2) kilometers for both distributions, though a slight imbalance of 19\% of trip length is estimated for real trips as compared to 17\% for synthesized trips. There is a steep decline of trips whose distances are beyond 5 kilometers in both real and synthetic representations. These statistical estimations are expected because travels within urban communities like in the case of our study region are relatively shorter than rural areas. The synthetic sequences present a near perfect fitting on trip lengths to the real sequences as shown in Figure \ref{fig:trip_len}c having an SRMSE of 0.211 and a correlation coefficient of 0.99 and an adjusted R$^2$ of 99\%.

\subsubsection{Route segment usage distribution}
The purpose of this metric is to evaluate the similarity in the frequency of trip routes taken by agents. While the model outputs sequences of trip destinations, we assume the shortest possible distance using the Dijkstra Algorithm \cite{johnson1973note} to derive the route itinerary from Montreal road network \cite{Montreal2020Road}. We compare the frequency of trip counts travelled on each route for both true and synthesized data. The efficiency of the synthesized trip sequences is evaluated by the similarity or equivalence in route usage counts observed on both true and synthetic trips. 

\begin{figure}[!h]
	\centering
	\includegraphics[width=0.5\textwidth]{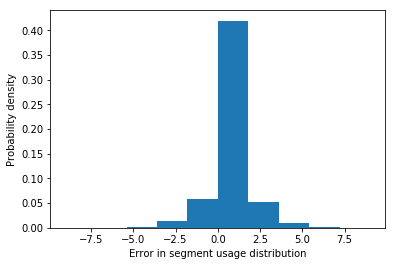}
	\caption{Distribution of differences in route segment usage for true and synthetic trips.}
	\label{fig:trip_hist}
\end{figure}

\begin{figure}[!h]
	\centering
	\includegraphics[width=1.00\textwidth]{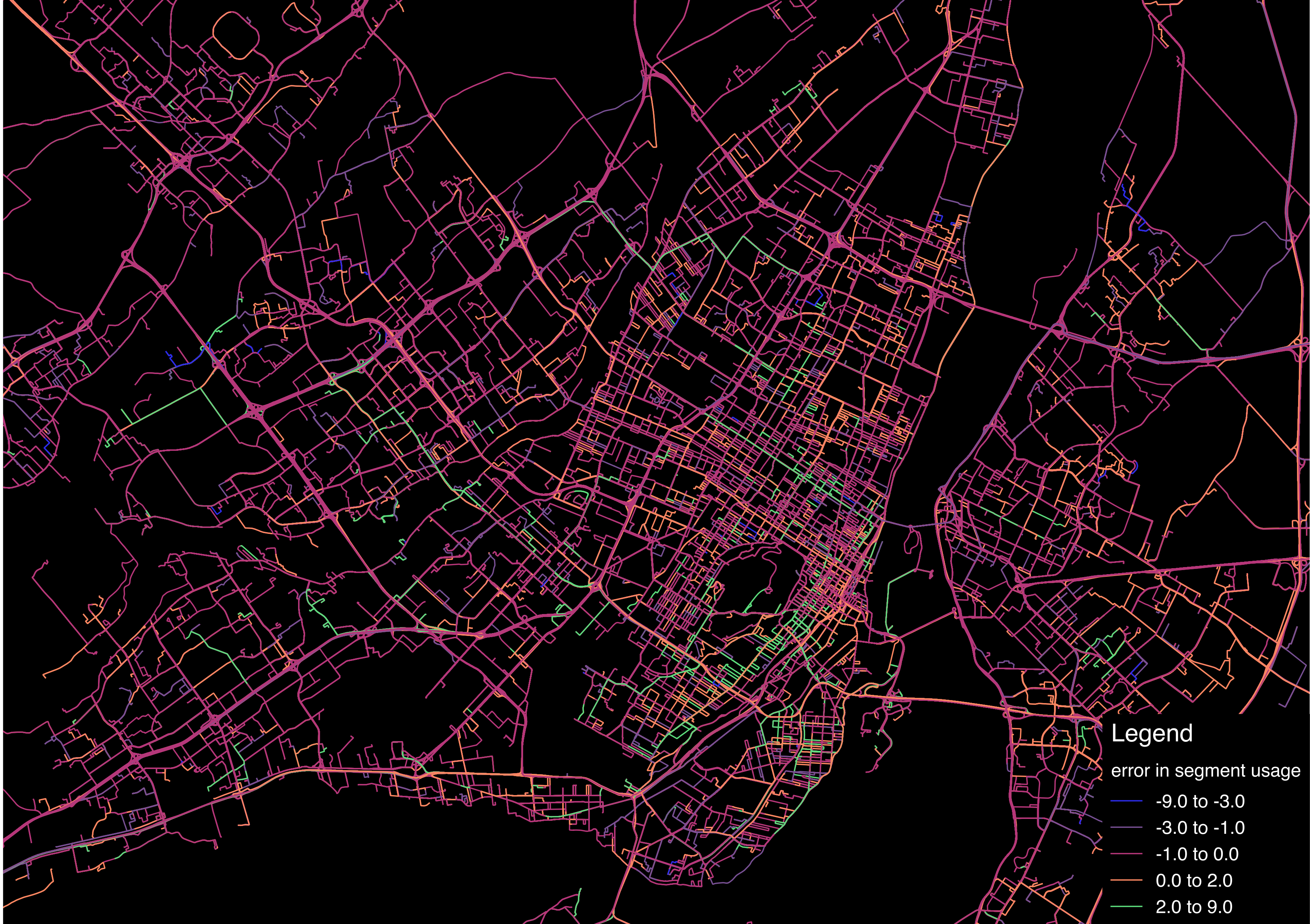}
	\caption{Route usage distribution of error in the simulated sequential trips of Greater Montreal Area}
	\label{fig:trip_map}
\end{figure}

In Figure \ref{fig:trip_hist}, a high proportion of routes show equivalent similarity on route usage for both true and synthesized trips. The model shows remarkable success in generating similar route usage frequencies at a probability density above 40\% recording the difference in usage counts between the true and synthesized. Route usage probabilities less than 5\% of the total routes exhibit variances in frequency within range of 1 to 5 counts symmetrically. We illustrate the error distribution of route usage for the Greater Montreal Area shown in Figure \ref{fig:trip_map}. A majority of the routes give a perfect fit of synthetic reconstruction marked by differences close to zero, colored in magenta on the route map.

\subsection{Sensitivity Analysis}
The aim of this analysis is to critically and systematically evaluate the performance, accuracy and elasticity of CTGAN for varying sample and categorical sizes when synthesizing individual level attributes or populations. The outputs are assessed using the Standard Root Mean Square Error, calculated by comparing the sample to the simulated population and the coefficient of determination, denoted by R$^2$. %The sensitivity analysis is categorized into two parts:

\subsubsection{Varying input sample size}
\begin{figure}[!h]
	\begin{subfigure}[t]{.45\textwidth}
		%\centering
		\includegraphics[width=\linewidth]{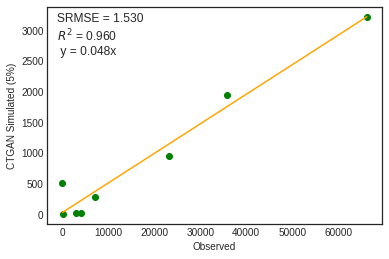}
		\caption{Sample size of \textbf{5\%}.}
	\end{subfigure}
	%\hfill
	\begin{subfigure}[t]{.45\textwidth}
		%\centering
		\includegraphics[width=\linewidth]{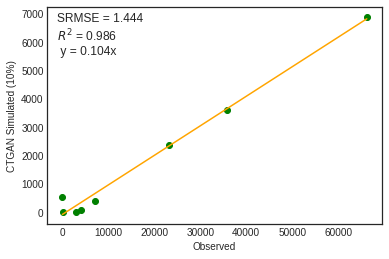}
		\caption{Sample size of \textbf{10\%}.}
	\end{subfigure}
	
	\medskip
	
	\begin{subfigure}[t]{.45\textwidth}
		%\centering
		\includegraphics[width=\linewidth]{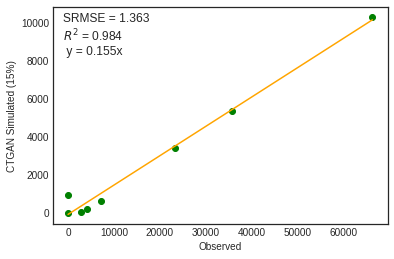}
		\caption{Sample size of \textbf{15\%}.}
	\end{subfigure}
	%\hfill
	\begin{subfigure}[t]{.45\textwidth}
		%\centering
		\includegraphics[width=\linewidth]{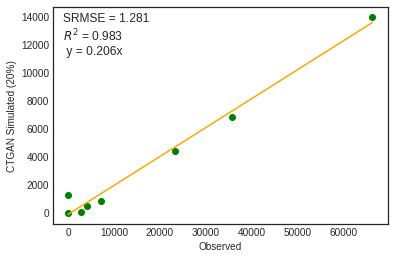}
		\caption{Sample size of \textbf{20\%}.}
	\end{subfigure}
	\caption{Uni-dimensional distribution of varying sampling sizes between observed and simulated observations.}
	\label{fig:uni_plot}
\end{figure}

In this approach, random samples are selected from the original sample with sizes of 5, 10, 15 and 20\%. The varying selected samples were independently trained as inputs to CTGAN. Scatter plots are shown in Figure \ref{fig:uni_plot} to depict the relationships between the observed and simulated for dimensions using different sampling sizes.  

With a sample size of 5\%, we observe a spread along the line of fit with an SRMSE of 1.530. Subsequently an improvement is observed as the sample size is increased to 10\% with declining SRMSE of 1.444. It is observed that the fit improves while minimizing spread when sample sizes are increased. This suggests the model performs better with an increase in sample size and smooths towards the distribution of the sample population with incremental sample ranges. Table \ref{tab:srmse} gives a summary of the performance for all simulated dimensions. As expected, a decline in the mean squared error for all synthesized dimensions is observed as sample sizes are increased.

\begin{table}[!h]
	\small
	\centering
	\begin{tabular}{|l|l|l|l|l|}
		\hline
		\textbf{Sample} & \textbf{Status} & \textbf{Gender} & \textbf{Permits} & \textbf{Age} \\ \hline
		5               & 1.530           & 0.950           & 1.762            & 1.202        \\ \hline
		10              & 1.444           & 0.900           & 1.666            & 1.123        \\ \hline
		15              & 1.363           & 0.850           & 1.575            & 1.052        \\ \hline
		20              & 1.281           & 0.800           & 1.465            & 0.994        \\ \hline
	\end{tabular}
	\caption{Standardized Root Mean Square Error (SRMSE) on varying samples of synthetic generation on varying sizes}.
	\label{tab:srmse}
\end{table}

\subsubsection{Inter-attribute relationships}
\begin{figure}[!h]
	\begin{subfigure}[t]{.45\textwidth}
		\centering
		\includegraphics[width=\linewidth]{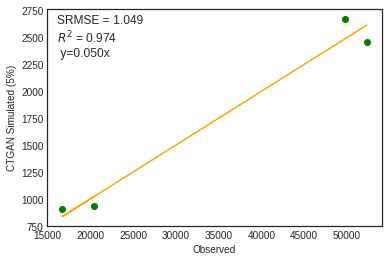}
		\caption{Sample size of \textbf{5\%}.}
	\end{subfigure}
	%\hfill
	\begin{subfigure}[t]{.45\textwidth}
		\centering
		\includegraphics[width=\linewidth]{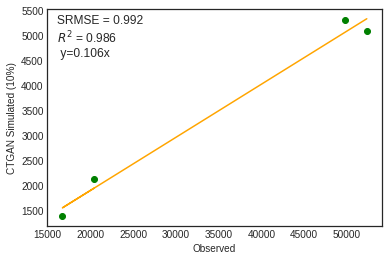}
		\caption{Sample size of \textbf{10\%}.}
	\end{subfigure}
	
	\medskip
	
	\begin{subfigure}[t]{.45\textwidth}
		\centering
		\includegraphics[width=\linewidth]{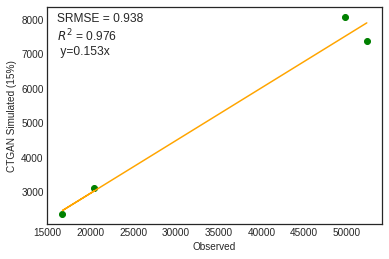}
		\caption{Sample size of \textbf{15\%}.}
	\end{subfigure}
	%\hfill
	\begin{subfigure}[t]{.45\textwidth}
		\centering
		\includegraphics[width=\linewidth]{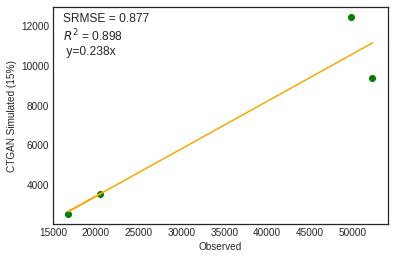}
		\caption{Sample size of \textbf{20\%}.}
	\end{subfigure}
	\caption{Conditional distributions for permit by gender between observed and simulated counts.}
	\label{fig:permit_gender_plot}
\end{figure}

\begin{figure}[!h]
	\begin{subfigure}[t]{.45\textwidth}
		\centering
		\includegraphics[width=\linewidth]{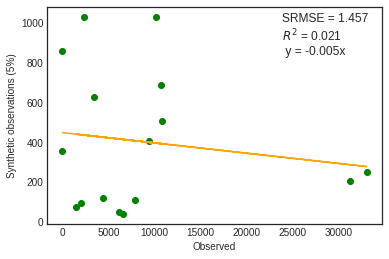}
		\caption{Sample size of \textbf{5\%}.}
	\end{subfigure}
	%\hfill
	\begin{subfigure}[t]{.45\textwidth}
		\centering
		\includegraphics[width=\linewidth]{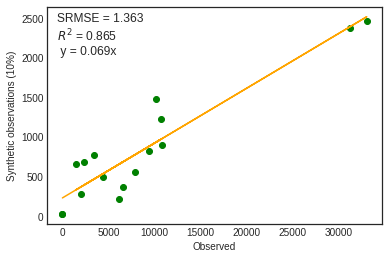}
		\caption{Sample size of \textbf{10\%}.}
	\end{subfigure}
	
	\medskip
	
	\begin{subfigure}[t]{.45\textwidth}
		%\centering
		\includegraphics[width=\linewidth]{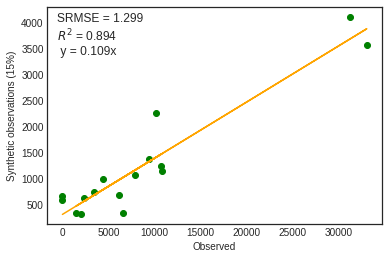}
		\caption{Sample size of \textbf{15\%}.}
	\end{subfigure}
	%\hfill
	\begin{subfigure}[t]{.45\textwidth}
		%\centering
		\includegraphics[width=\linewidth]{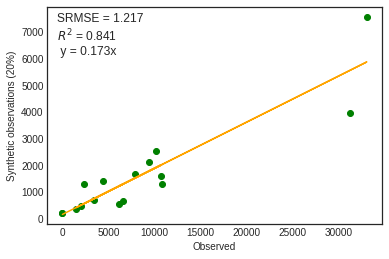}
		\caption{Sample size of \textbf{20\%}.}
	\end{subfigure}
	\caption{Full joint distributions for all variables between observed and simulated counts.}
	\label{fig:joint_plot}
\end{figure}

This analysis considers how well the synthetic model recreates the observed relationships between attributes in the original sample population for varying sample sizes (i.e. 5, 10, 15, 20\%). The results in Figure \ref{fig:permit_gender_plot} show the performance of the conditional probabilities for Permit by Gender attributes and Age Group by Gender attributes. The line of fit exhibits a balance population between counts of the conditionals. As observed, the increase in sample sizes reduces the mean square errors from 1.049 for a 5\% sample size to SRMSE of 0.992 for a 10\% sample size, these steadily decline in SRMSE values for increasing sample sizes. This suggests the model improves on learning a fit of the conditional distributions between attributes and subsequently smooths the distribution of the increasing sample sizes toward the distribution of the sample population. Similarly, we evaluated the full joint distribution for all variables between the sample population and synthesized population. The output observations were re-sampled and evaluated. Using a 5\% sample size as shown in Figure \ref{fig:joint_plot}, there is a wider distribution spread between observed and synthetic of SRMSE at 1.457, while the line of fit shows a spread of points along it. This suggests an imbalance in the population summaries between observed and synthesized observations with a weaker distribution fit compared to the sample population depicted by the spread. It can be seen from the analysis that the model shows consistency in learning the inter-attributes relationships and full joint distributions between all attributes when the sample sizes are increased.

\subsubsection{Varying categorical sizes}
\begin{figure}[!h]
	\begin{subfigure}[t]{.45\textwidth}
		%\centering
		\includegraphics[width=0.95\linewidth]{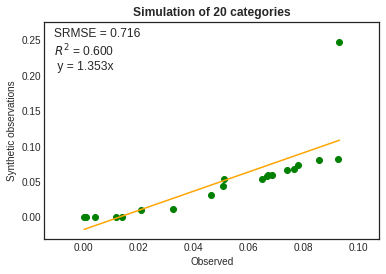}
		\caption{Simulation for category size of \textbf{20}.}
	\end{subfigure}
	%\hfill
	\begin{subfigure}[t]{.45\textwidth}
		%\centering
		\includegraphics[width=0.95\linewidth]{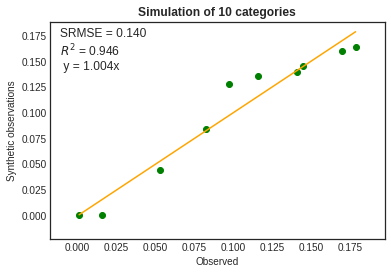}
		\caption{Simulation for category size of \textbf{10}.}
	\end{subfigure}
	
	\medskip
	
	\begin{subfigure}[t]{.45\textwidth}
		%\centering
		\includegraphics[width=0.95\linewidth]{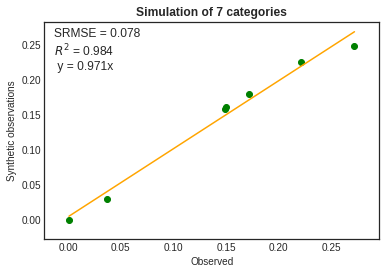}
		\caption{Simulation for category size of \textbf{7}.}
	\end{subfigure}
	%\hfill
	\begin{subfigure}[t]{.45\textwidth}
		%\centering
		\includegraphics[width=0.95\linewidth]{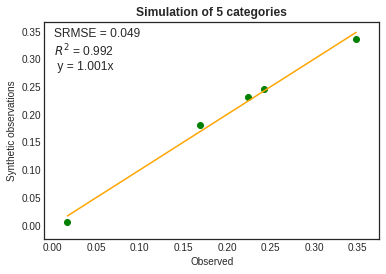}
		\caption{Simulation for category size of \textbf{5}.}
	\end{subfigure}
	\caption{Distribution of varying categorical sizes (age discretized).}
	\label{fig:varying_plot}
\end{figure}

In the final experiment, we evaluate the performance of CTGAN for varying categorical sizes. For this purpose, the attribute ``Age" is converted from numerical to categorical input and subsequently discretized into bin sizes of 5, 10, 15 and 20 categories of age groups. The model is retrained with the discretized categories and the output is represented in Figure \ref{fig:varying_plot}. At a category size of 20, we observe a weaker correlation along the line of fit suggesting an imbalance between population counts of observed and simulated observations having a high SRMSE of 0.716. The output of trained samples on category size of 10 shows a better improvement of fit with a wide spread along its perfect line of fit. We observe a sequential improvement with a reduction to size of categories for 7 and 5 categories. This suggest the model is able to smoothen the distributions of minimal categories or modes. This could have arisen because of the lack of diversity/mode dropping and non-convergence that is notable limitation in GANs \cite{kodali2017convergence, arora2017gans}.

\section{Discussions and conclusions}
A novel deep learning generative model for reconstructing synthetic agents having tabular and sequential location-based travel information is presented. Specifically, we combine two generators and two discriminators to design the Composite Travel GAN (CTGAN) architecture that outputs both tabular and sequential attributes simultaneously. The work compared the statistical similarities of the synthetic tabular results of the CTGAN with synthetic results from the VAE. The models were tested with sample population data from the origin-destination survey of the region of Greater Montreal (Canada) in 2013. The CTGAN outperformed the VAE in terms of synthetic generation of tabular data. 

Our results show the capability and success of CTGAN to recreate the marginals of attributes for both the tabular and sequential samples while maintaining inter-attribute relationships. We observed improvement of the performance of the model through scaling of different sample sizes with a better output for the large sample sizes that smoothens the learning distribution to the underlying distribution of the sample population. Sampling variation has a significant impact on the representation of the attributes and inter-attributes relations as evident in the analysis of the varying sizes. Based on this trend, it can be concluded that the model will perform better when a larger sample population is provided.

When implementing CTGAN, we observed the following drawbacks. There was significantly longer training time to synthesizing both tabular and sequential data simultaneously. 12 hours were required to train and synthesize 100,000 simulated household samples. Also, CTGAN showed difficulty in training sequences of more than 5000 complete trips hence samples had to be batched for training. These drawbacks limit the adaptation of the model on real travel datasets which could have millions of travel records. In this regard, future work will consider deploying the model in a distributed computing framework and parallelized training on multiple nodes to improve on the training time and increase capacity for optimal model training. We also seek to consider improving the generative framework with losses to control the level of privacy that can be achieved. We will be able to control the expected privacy, especially in cases of releasing data to non-trusted data agents. While this paper is one of the first studies using generative models on travel data, we plan to explore methods that will be needed to improve the utility and privacy of the models when publicly releasing the synthetic datasets. We will work to extend this research on the generation of synthesized continuous mobility trajectories. We will explore the use of federated learning and Blockchain for Smart Mobility Data-markets (BSMD) framework proposed by \cite{lopez2020multi} to estimate CTGAN without directly accessing the sample, which may result in compromising the privacy of the individuals in the sample.

\section*{Acknowledgments}

The research presented in this paper has been funded by the following sources: Social Sciences and Humanities Research Council, Canada (890-2015-0022), Canada Research Chairs (950- 224364).


\newpage
\appendix

\section{Study Area}

\begin{figure}[!h]
	\centering
	\includegraphics[width=1.00\textwidth]{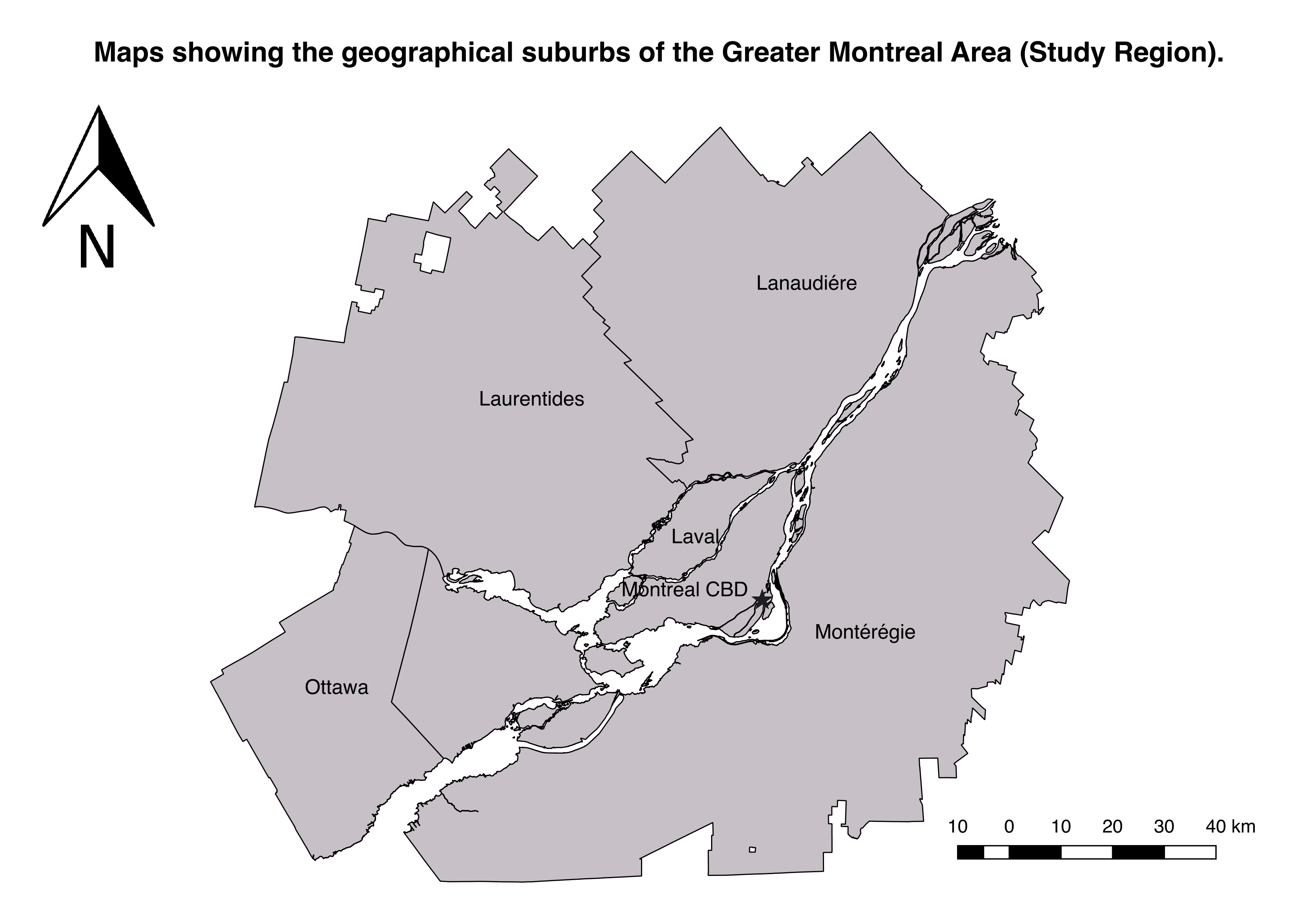}
	\caption{Map of geographic areas of the Greater Montreal Area}
	\label{fig:study_area}
\end{figure}

The map in Figure \ref{fig:study_area} shows the geographical extent of the study area. The map states the boundaries of the census metropolitan areas  within the Greater Montreal Area.

\newpage
\section{Data Preparation}

In this section, we discuss the procedures that were adopted to prepare the data for the generative modelling. The trip data for this project composed on numerical, categorical and location geographic variables as well as location sequences.

\subsection{Numerical attributes}
The objective of processing numerical attributes is to normalize and scale within range of -1 to 1. The approaches of Scaling and Normalization standardizes numeric inputs into data points that are suitable for Neural Networks. Standardizing data points transform that into a resulting distribution with a mean of 0 and a standard deviation of 1. Normalization is defined by:

\begin{equation} \label{eq5}
x^{|} = \frac{x-x_{mean}}{x_{max}-x_{min}}
\end{equation}

where \textit{X} is the feature vector, \textit{X$_{mean}$} is the mean of the feature vector, \textit{X$_{min}$} is the minimum of the feature vector and \textit{X$_{max}$} is the maximum of the feature vector. We implemented the normalization using the Scikit-learn \cite{mcginnis2018category} Pre-processing framework available in Python. The package presents two libraries: MinMaxScaler and StandardScaler. The MinMaxScaler library normalizes a feature to range of 0 to 1 while the StandardScaler library standardizes the data points to a mean of 0.

\subsection{Categorical attributes}
When processing categorical attributes, we consider two categories namely low and high cardinality. Low cardinality refers to variables with a minimum of 20 unique variables while High cardinality referes to variables with 20 or more unique variables. For low cardinal variables, we apply the one-hot encoding technique. One-hot encoding \cite{orup1999fly} converts categorical variables to binary combinations of values with a single high (1) bit and all the others low (0). This encoding technique derives an integer representation for category values with a length of the encoded vectors equivalent to the number of unique values of the variable. This technique becomes inefficient when implemented on high categorical values since larger matrices are created with a drawback on computation. On the other hand, we employ feature embeddings \cite{wang2015word, guo2016entity} to encode high cardinal values to fixed dimensional real values. Feature embeddings derive unique real-valued vectors to represent each category. We employ Keras layer embeddings for generation of feature embeddings for high cardinal categories.

\subsection{Route Itinerary}

For the purposes of trip sequences, the model demand complete route itineraries between origin and destination geographic points. The travel routes were generated with the shortest distance path between an origin and a destination data points. The Open Source Routing Machine (OSRM) allows a public accessible Application Programming Interface (API) available at http://project-osrm.org. The API endpoint returns a sequence of geographic points stating the complete geographical route itinerary.


%% References



%% New version of the num-names style
\bibliographystyle{abbrv}
\bibliography{references.bib}
\end{document}